\documentclass[runningheads]{llncs}
\usepackage{graphicx}

\usepackage{tikz}
\usepackage{comment}
\usepackage{amsmath,amssymb} 
\usepackage{color}

\usepackage{graphicx}
\usepackage{amssymb}
\usepackage{booktabs}
\usepackage{framed}
\usepackage{amssymb}
\usepackage{amsfonts}
\usepackage{mathrsfs}
\usepackage{mathtools}
\usepackage{array}
\usepackage{verbatim}
\usepackage{enumerate}
\usepackage{bbm}
\usepackage{commath}
\usepackage{wrapfig}
\usepackage{amsbsy}
\usepackage{float}
\usepackage{tabularx}
\usepackage{listings}
\usepackage{xcolor}
\usepackage{enumitem}
\usepackage{multirow}
\usepackage{booktabs}
\usepackage{stmaryrd}
\usepackage{algorithm}
\usepackage{algorithmic}
\usepackage{grffile}
\usepackage{orcidlink} 
\usepackage[misc]{ifsym}

\begin{document}
\pagestyle{headings}
\mainmatter
\def\ECCVSubNumber{4800}  

\title{DLME: Deep Local-flatness Manifold Embedding} 

\titlerunning{DLME: Deep Local-Flatness Manifold Embedding}
%
\author{
  Zelin Zang\inst{1,2}$^{\divideontimes}$\orcidlink{0000-0003-2831-5437} \and
  Siyuan Li\inst{1,2}$^{\divideontimes}$\orcidlink{0000-0001-6806-2468} \and
  Di Wu\inst{1,2} \and
  Ge Wang\inst{1,2} \and
  Kai Wang\inst{3} \and\\
  Lei Shang\inst{3} \and
  Baigui Sun\inst{3} \and
  Hao Li\inst{3} \and 
  Stan Z. Li\inst{1,2,}$^{\textrm{\Letter}}$\orcidlink{0000-0002-2961-8096}
  }

\authorrunning{Zelin Zang et al.}
%
  
\institute{ 
  Zhejiang University, Hangzhou, 310000, China
  \email{\texttt{\{zangzelin,stan.zq.li\}@westlake.edu.cn}}
  \and
  Westlake University, AI Lab, School of Engineering, Hangzhou, 310000, China\\
  \and
  Alibaba Group, Hangzhou, China\\
  {${\divideontimes}$ Equal contribution, \textrm{\Letter} Corresponding author}
  }

\maketitle

\begin{abstract}
  Manifold learning~(ML) aims to seek low-dimensional embedding from high-dimensional data. The problem is challenging on real-world datasets, especially with under-sampling data, and we find that previous methods perform poorly in this case. Generally, ML methods first transform input data into a low-dimensional embedding space to maintain the data's geometric structure and subsequently perform downstream tasks therein. The poor local connectivity of under-sampling data in the former step and inappropriate optimization objectives in the latter step leads to two problems: \emph{structural distortion} and \emph{underconstrained embedding}. This paper proposes a novel ML framework named Deep Local-flatness Manifold Embedding (DLME) to solve these problems. The proposed DLME constructs semantic manifolds by data augmentation and overcomes the {structural distortion} problem using a smoothness constrained based on a \emph{local flatness} assumption about the manifold. To overcome the {underconstrained embedding} problem, we design a loss and theoretically demonstrate that it leads to a more suitable embedding based on the local flatness. Experiments on three types of datasets (toy, biological, and image) for various downstream tasks (classification, clustering, and visualization) show that our proposed DLME outperforms state-of-the-art ML and contrastive learning methods.
\end{abstract}

\section{Introduction}
\label{sec_intro}

\begin{figure}[t]
  \centering
  \includegraphics[width=4.7in]{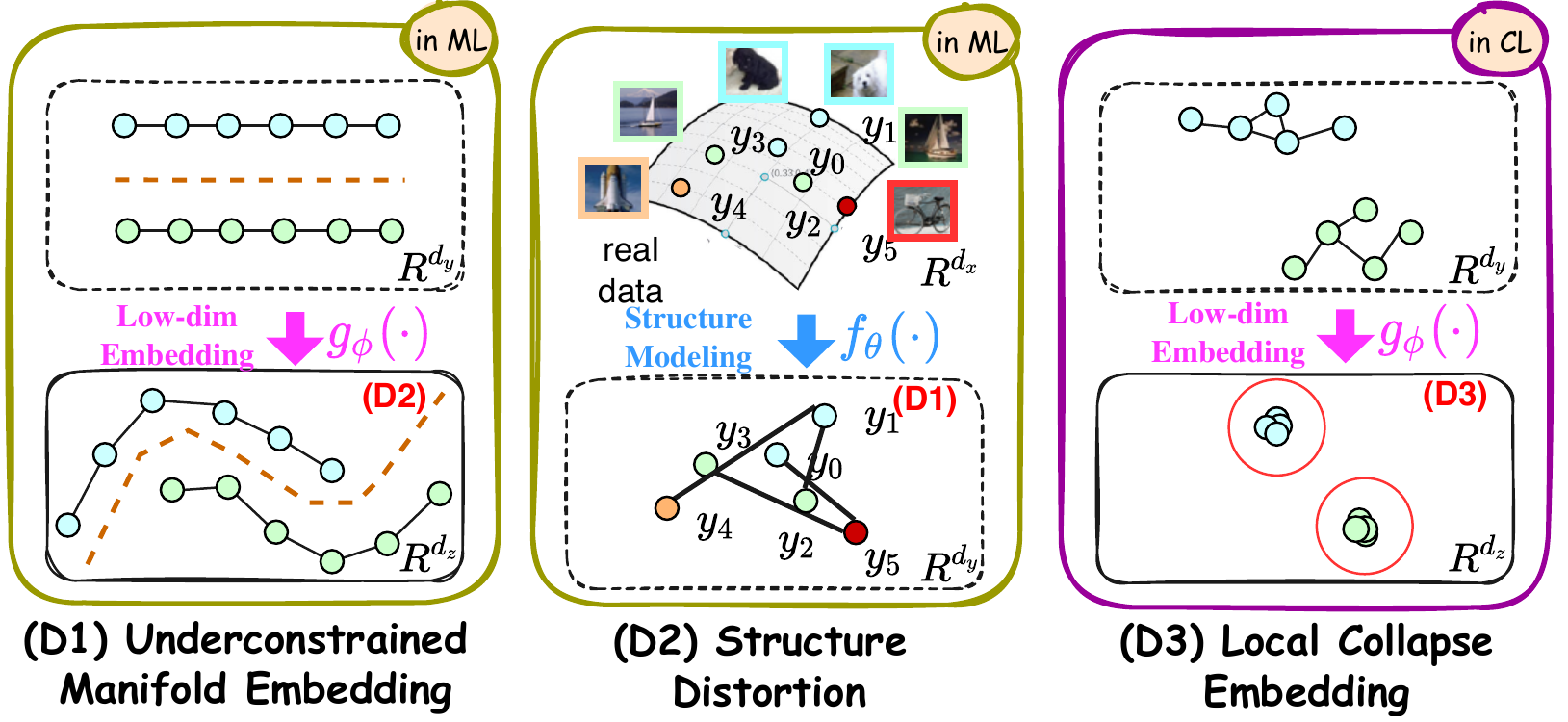}
  \caption{
    Problems in ML and CL.
    \textbf{(D1)} 
    Local field of view
    \&
    first-order (similarity/dissimilarity) constraints
    $\rightarrow$
    underconstrained manifold embedding.
    \textbf{(D2)} 
    complexity of real-world data (ultra-high dimensionality or non well-sampling)
    $\rightarrow$
    broke the local connectivity of manifold
    $\rightarrow$
    structure distortions.
    \textbf{(D3)} 
    unsmoothed losses function
    $\rightarrow$
    local collapse embedding.
  }
  \label{fig:current_problem}
\end{figure}

The intrinsic dimension of high-dimensional data is usually much lower and how to effectively learn a low-dimensional representation is a fundamental problem in traditional machine learning~\cite{pers2009validation}, data mining~\cite{agrawal1998automatic}, and pattern recognition~\cite{donoho2000high}. Manifold learning~(ML), based on solid theoretical foundations and assumptions, discusses manifold representation problems under unsupervised conditions and has a far-reaching impact.
However, practical applications of the manifold learning method are limited in real-world scenarios, and we attribute the reasons to the following two reasons. 
\textbf{(D1) Underconstrained manifold embedding.} ML methods focus on local relationships, while it is prone to distorted embeddings that affect the performance of downstream tasks (in Fig.~\ref{fig:current_problem}~(D2) and Fig.~\ref{fig:swishrollandbell}). Paper~\cite{kobak2019umap,kobak2021initialization} suggests even the most advanced ML methods lose performance on downstream tasks due to inadequate constraints on the latent space. The reason is attributed to the limitations of traditional ML methods based on similarity/dissimilarity loss function design.
\textbf{(D2) Structural distortion.} ML methods focus on handcraft or easy datasets and are not satisfactory in handling real-world datasets. Most of these approaches use the locally connected graphs constructed in the input space to define structure-preserving unsupervised learning loss functions~\cite{mcinnes_umap_2018,maaten_visualizing_2008}. These methods introduce a stringent assumption~(local connectivity assumption~(LCA)) which suggests the metric of input data well describes the data's neighbor relationship. However, LCA requires the data to be densely sampled and too ideal in the real world, e.g., pictures of two dogs are not necessarily similar in terms of pixel metric~(in Fig.~\ref{fig:current_problem}~(D2)).

\begin{figure*}[t]
  \centering
  \includegraphics[width=4.5in]{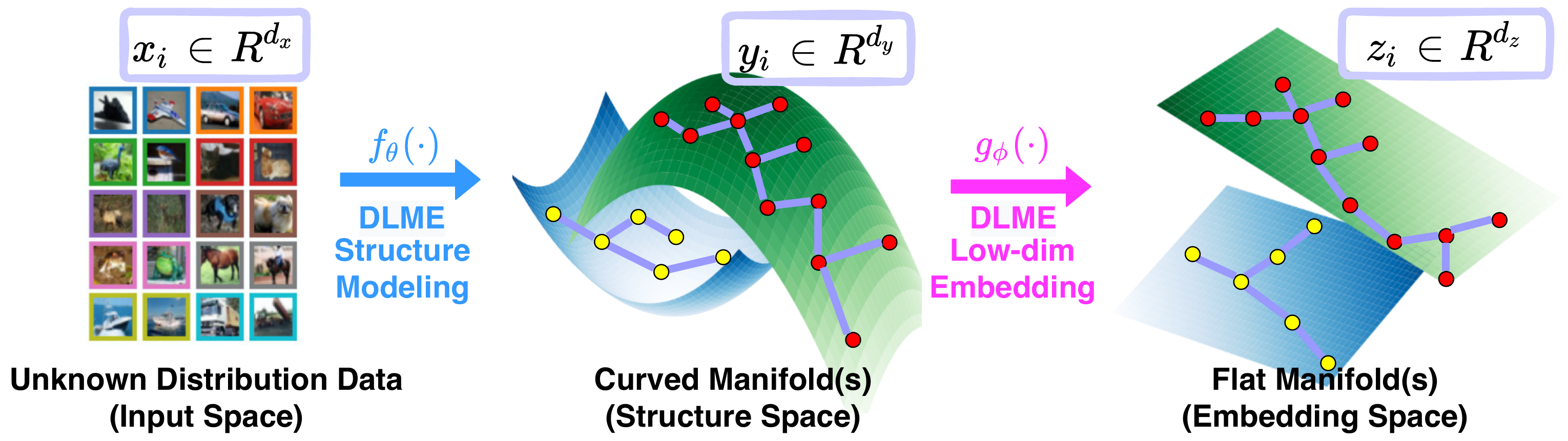}
  \caption{
    DLME includes the structure modeling network $f_{\theta}(\cdot)$ and low-dim embedding network $g_{\phi}(\cdot)$. 
    The $f_{\theta}(\cdot)$ maps the input data into structure space to describe the data relationship.
    The $g_{\phi}(\cdot)$ maps the curled manifold into the flat embedding space to improve the discriminative performance and friendliness to downstream tasks. $f_{\theta}(\cdot)$ and $g_{\phi}(\cdot)$ are compatible with any neural network.
  }
  \label{fig_intro}
\end{figure*}

Meanwhile, Contrastive Learning~(CL) is enthusiastically discussed in the image and NLP fields~\cite{chen_simple_2020,he_momentum_2020,grill_bootstrap_2020}. 
These methods have shown excellent performance by introducing prior knowledge of the data with the help of data augmentation. However, we have encountered significant difficulties applying such techniques to the ML domain. The above methods require a large amount of data for pre-training~\cite{weng2021contrastive,zbontar2021barlow}, so it is not easy to achieve good results in areas where data is expensive (e.g., biology, medicine, etc.).
We consider that the core issues can be summarized as \textbf{(D3) Local collapse embedding.} The unsmoothed loss of the CL leads to the model that is prone to local collapse and requires a large diversity of data to learn valid knowledge~(in Fig.~\ref{fig_intro}~(D3)).

We want to propose a novel deep ML model to constrain the latent space better and solve the structural distortion problem with the help of CL. At the same time, we hope the proposed method avoids the local collapse phenomenon in the CL. The process of ML perspective includes \textit{structural modeling sub-processes} and \textit{low-dimensional embedding sub-processes}. The structural modeling sub-process obtains the graph structures of data manifolds by measuring the relationship of each sample pair which serves as the guidance for low-dimensional embedding. The low-dimensional embedding process maps the provided graph structures into the embedding space.

We propose a novel deep ML framework named deep local-flatness manifold embedding (DLME) to solve both problems by merging the advantages of ML and CL.
Firstly, a novel local flatness assumption~(LFA) is proposed to obtain a reasonable latent space by adding a second-order manifold curvature constraint (for \textbf{D1}), thus improving the performance on downstream tasks.
Secondly, a new neural network framework is designed to accommodate data augmentation and emhance the net work trainging~(for \textbf{D2}). DLME framework accomplishes the two sub-processes with two networks ($f_\theta(\cdot)$ and $g_\phi(\cdot)$) optimized by the proposed DLME loss between the latent space of $f_\theta(\cdot)$ and $g_\phi(\cdot)$ in an end-to-end manner~(framework is in Fig.~\ref{fig_intro}).
Furthermore, an LFA-based smoother loss function is designed to accommodate data augmentation. It is based on a long-tailed t-distribution and guides the network learning through the two latent spaces (for \textbf{D3}). Finally, we further illustrate mathematically: (1) the differences between DLME loss and conventional CL loss and (2) why DLME loss can obtain a locally flat embedding.

In short, DLME makes the following contributions:
  (1) DLME provides a novel deep ML framework that utilizes neural networks instead of distance metrics in data space to better model structural relationships, thus overcoming structural distortions.
  (2) DLME put forward the concept of local flatness and theoretically discusses that the DLME loss can enhance the flatness of manifolds and obtain transfer abilities to downstream tasks.
  (3) The effectiveness of DLME is demonstrated on three downstream tasks with three types of datasets (toy, biological, and image). Experiment results show that DLME outperforms current state-of-the-art ML and CL methods.

\section{Related Works}
\label{sec_related}

In \textbf{manifold learning}, MDS~\cite{kruskal1964nonmetric}, ISOMAP~\cite{tenenbaum_global_2000}, and LLE~\cite{roweis_nonlinear_2000} model the structure of data manifolds based on local or global distances (dissimilarity) and linearly project to the low-dimensional space. SNE~\cite{hinton_stochastic_2003}, t-SNE~\cite{maaten_visualizing_2008} and UMAP~\cite{mcinnes_umap_2018} use  normal distribution to define local similarities and apply the Gaussian or t-distribution kernel to transform the distance into the pair-wise similarity for \textit{structural modeling}. They perform the manifold embedding by preserving the local geometric structures explored from the input data.

In \textbf{deep manifold learning}, Parametric t-SNE (P-TSNE)~\cite{maaten_learning_2009} and Parametric UMAP~\cite{sainburg_parametric_2021} learn more complex manifolds by non-linear neural networks and can transfer to unseen data. However, they inherit the original \textit{structural modeling} strategies in t-SNE and UMAP. Topological autoencoder (TAE), Geometry Regularized AutoEncoders (GRAE)~\cite{Duque2020GRAE}, and ivis~\cite{szubert_structure_preserving_2019} abandon the accurate modeling of input data and directly achieve \textit{low-dimensional embedding} using distances or contrast training. 

In \textbf{self-supervised contrastive learning}, contrastive-based methods~\cite{wu_unsupervised_2018,chen_simple_2020,he_momentum_2020,grill_bootstrap_2020} which learns instance-level discriminative representations by contrasting positive and negative views have largely reduced the performance gap between supervised models on various downstream tasks. 
Deep clustering methods
is another popular form of self-supervised pertraining.
\section{Methods}
\label{sec_method}

\subsection{Problem Description and Local Flatness Assumptions}
According to Nash embedding theorems~\cite{nash1956imbedding}, we mainly discuss manifolds represented in Euclidean coordinates and provide the definitions of manifold learning (ML) and deep manifold learning (DML) in practical scenarios.

\noindent \textbf{Definition 1 (Manifold Learning, ML)}. 
Let $\mathcal{M}$ be a $d$-dimensional embedding in Euclidean space $\mathbb{R}^d$ and $f: \mathcal{M} \rightarrow \mathbb{R}^{D}$ be a diffeomorphic embedding map, for $D>d$, the purpose of manifold learning is to find $\{z_i\}_{i=1}^N, z_i \in \mathcal{M}$ from the sufficient sampled(observed) data $ X=\{x_i\}_{i=1}^N, x_i \in \mathbb{R}^D$.

Based on Definition 1, the DML aims finding the embedding $\{z_i\}_{i=1}^N, z_i \in \mathcal{M}$ by mapping $g_{\theta}: \mathbb{R}^{D}\rightarrow \mathbb{R}^{d}$ with the neural network parameters $\theta$. Each ML method designs a loss function based on the specific manifold assumption to map the observed data $\{x_i\}$ back to the intrinsic manifold $\{z_i\}$. For example, LLE~\cite{roweis2000nonlinear} assumes that the local manifold is linear, and UMAP~\cite{mcinnes_umap_2018} assumes that the local manifold is uniform. We propose a novel assumption, considering the nature of the manifold is local flatness.

\noindent \textbf{Assumptions 1 (Local Flatness Assumption, LFA)}. 
Let $\mathcal{M}$ be a manifold, and $\{x_i\}$ be a set of observations in the manifold. 
We expect each data point and its neighbors to lie on or close to a local flatness patch. 
The mean curvature $\overline{K}_\mathcal{M}$ is introduced to quantify the flatness of high-dimensional manifolds according to the Gauss-Bonnet theorem~\cite{satake1957gauss},
\begin{equation}
  \begin{aligned}
    \overline{K}_\mathcal{M} = &\sum_{x_i \in X} k(x_i)\\
    k(x_i) = 
      & 2 \pi \chi(\mathcal{M}) 
      - 
      \theta{(x_{|\text{H}_1(x_i)|}, x_i, x_0)}
      -\sum_{
        \substack{
          j \in \{ 0, 1, \cdots, |\text{H}_1(x_i)|-1 \}
        } 
      }
      \theta{(x_j, x_i, x_{j+1})},
  \end{aligned}
  \label{eq:curvature}
\end{equation}
where $\mathcal{\chi(\mathcal{M})}$ is Euler Characteristic~\cite{harer1986euler}. The $\text{H}_1(x_i)$ is the hop-1 neighbor of $x_i$, and $\theta{(x_i, x_j, x_{k})}$ is the angle of three point $x_i$ $x_j$, $x_k$. 

\textbf{From first-order constraint to second-order curvature constraint}. ML methods (e.g., LLE and UMAP) design distance-preserving or similarity-preserving objective functions, hoping to guarantee the first-order relationship of the data. However, first-order relation preservation is not a tight enough constraint if the local structure of the manifold is simple, thus leading to underconstrained manifold embedding in ML.
We introduce a second-order~(curvature) constraint to solve the distortion problem. Due to the expensive complexity of second-order losses, we directly minimize the manifold's curvature by a mundane flatness assumption. 

\textbf{The Empirical benefits of LFA.}
Similar to most ML and CL methods, LFA is an assumption of latent space, which is beneficial for downstream tasks. In the case of the single-manifold, the assumption of `Local Flatness' reduces curling in the unsuitable embedding space (see Fig.~\ref{fig:swishrollandbell}), thus avoiding distortion during embedding. In the case of the multi-manifolds, assuming `Local Flatness' can simplify the discriminative relations of multi-manifolds. Therefore, the proposed assumption can avoid representation collapse. Meanwhile, it also reduces the possibility of different manifolds overlapping so that the downstream can be accomplished by a simple linear model easily.

\subsection{DLME Framework}
\label{subsec_DLME}
\begin{figure}[t]
  \centering
  \includegraphics[width=2.2in]{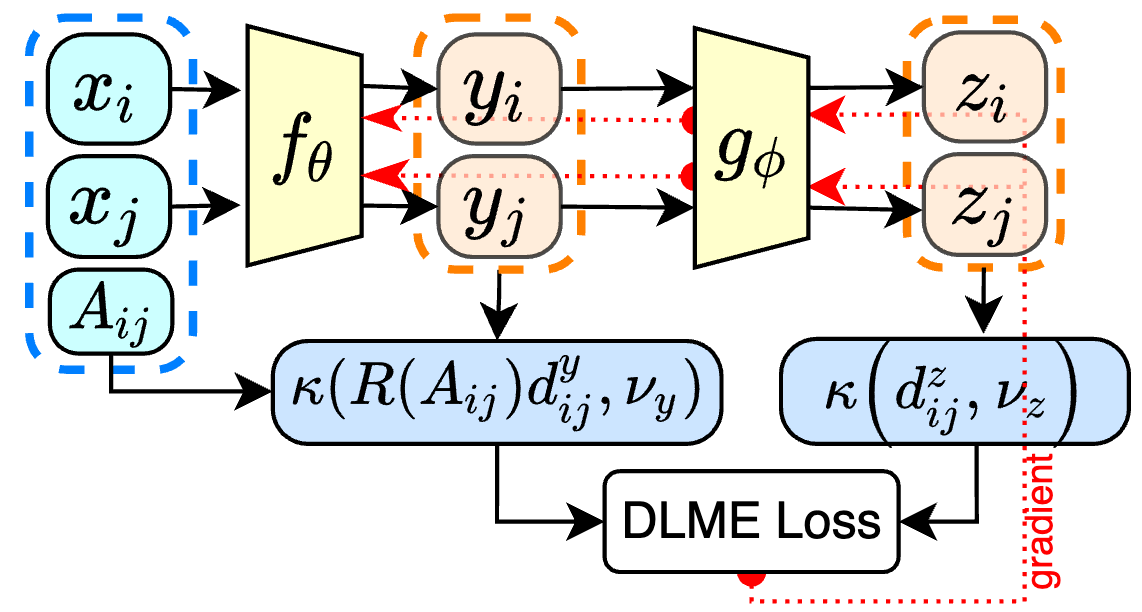}
  \caption{
    The framework of DLME. $(x_i,x_j)$ is a pair of input sample, and the neighbor relationship $A_{ij}$ indicates whether $x_i$ and $x_j$ are homologous pairs. The red dashed line marks the direction of the gradient back-propagation.
  }
  \label{fig:Fig-structure}
\end{figure}

As shown in in Fig.~\ref{fig:Fig-structure}, the DLME framework contains two neural networks ($f_{\theta}$ and $g_{\phi}$) and a DLME loss function $L_D$. The network $f_{\theta}$ achieves \textit{structural modeling} in its structure space $\mathbb{R}^{d_y}$, and the network $g_{\phi}$ learns \textit{low-dimensional embedding} in the embedding space $\mathbb{R}^{d_z}$. The DLME loss is calculated based on the $A_{ij}$ and the pairwise similarity in spaces $\mathbb{R}^{d_y}$ and $\mathbb{R}^{d_z}$ used to train two neural networks from scratch. The $A_{ij}$ indicate the homologous relationships, if $x_i$ and $x_j$ are augmentations of the same original data, then $A_{ij}=1$ else $A_{ij}=0$. 

\textbf{Data augmentation for solving structural distortion.}
ML methods have difficulty efficiently identifying neighboring nodes, causing structural distortions, when dealing with complex and not well-sampled data. DLME solves this problem with a priori knowledge provided by data augmentation. Data augmentation schemes have been widely used in self-supervised contrastive learning~(CL) to solve problems in CV and NLP. From the ML perspective, data augmentation is a technique to make new observations in the intrinsic manifold based on prior knowledge. Since data augmentation changes the semantics of the original data as little as possible, it generates specific neighborhood data for each isolated data when the local connectivity of ML data is broken. DLME trains a neural network $f_\theta(\cdot)$. The $f_\theta(\cdot)$ is guided by data augmentation and loss functions to map the data into a latent space that better guarantees local connectivity.

Data augmentation is designed based on domain knowledge. For example, in CV datasets, operations such as color jittering~\cite{zhan2020online}, random cropping~\cite{cheng2020random}, applying Gaussian blur~\cite{flusser2015recognition}, Mixup~\cite{liu2021automix,li2021boosting,liu2022decoupled} are proven useful. In biology and some easy datasets, linear combinations $\tau_{lc}(\cdot)$ in k-nearest neighbor data is a simple and effective way. The linear combinations is 
\begin{equation}
  \tau_{lc}(x) = r x  + (1-r) x^n, x^n \sim \text{KNN}(x),
\end{equation}
where $x^n$ is sampled from the neighborhood set of data $x$, and $r\in[0,1]$ is a combination parameter. For special domain data, the prior knowledge in the domain can be used to establish data augmentation.

\textbf{The forward propagation} of DLME is,
\begin{equation}
  \begin{aligned}
    y_i &= f_{\theta}(x_i), y_i \in \mathbb{R}^{d_y}, x_i \sim \tau(x), x_j \sim \tau(x),\\
    z_i &= g_{\phi}(y_i), z_i \in \mathbb{R}^{d_z}, d_z < d_y,
  \end{aligned}
\end{equation}
where $x_i$ and $x_j$ sampled form different random augmentation of raw data $x$, the ${d_y}$ and ${d_z}$ are the dimension number of $\mathbb{R}^{d_y}$ and $\mathbb{R}^{d_z}$.

\textbf{The loss function} of DLME is,
\begin{equation}
  \begin{aligned}
    L_{\text{D}} & \!=\! E_{
      \substack{ 
          x_i, x_j   
          }    
      } 
      \left[ 
        \mathcal{D}\left(
           \kappa \left(
              R(A_{ij}) d^y_{ij}, \nu_y 
              \right)
          , \kappa \left( d^z_{ij},\nu_z \right)
           \right)
      \right], \\
  \end{aligned}
\label{eq:final_model}
\end{equation}
where $d^y_{ij} = d(y_i, y_j)$, $d^z_{ij} = d(z_i, z_j)$ and $d^y_{ij}$, $d^z_{ij}$ are the distance metrics of data node $i$ and $j$ in spaces $\mathbb{R}^{d_y}$ and $\mathbb{R}^{d_z}$. 
The two-way divergence \cite{li2020deep} $\mathcal{D}(p, q)$ is introduced to measure the dis-similarity between two latent spaces,
\begin{equation}
    \mathcal{D}\left(p, q\right)  = p \log q + (1-p)\log(1-q),
    \label{eq:D}
\end{equation}
where $p \in [0, 1]$. Notice that $\mathcal{D}(p, q)$ is a continuous version of the cross-entropy loss. The two-way divergence is used to guide the pairwise similarity of two latent spaces to fit each other. The effect of the loss function on the two networks will be discussed in Sec~\ref{subsec_structural} and Sec~\ref{sub_flateness}.

The structure space requires a larger dimensionality to accurately measure data relationships, while the embedding space requires sufficient compression of the output dimension. Thus the $t$-distribution kernel function is used to calculate the pairwise similarity. The different degrees of freedom $\nu_y$ and $\nu_z$ in different spaces are essential to enhance the flatness of embedding space~(in Sec~\ref{sub_flateness}). 
\begin{equation}
    \kappa \left( d, \nu \right) 
    =
    \frac
    {\operatorname{Gam}\left(\frac{\nu +1}{2}\right)}
    {\sqrt{\nu  \pi} \operatorname{Gam}\left(\frac{\nu }{2}\right)}
    \left(
      1+\frac{d^{2}}{\nu }
    \right)^{-\frac{\nu +1}{2}},
    \label{eq:t_dis}
\end{equation}
where $\operatorname{Gam}(\cdot)$ is the Gamma function, and the degrees of freedom $\nu$ controls the shape of the kernel function. 

DLME design $R(A_{ij})$ to integrate the neighborhood information in $A_{ij}$. 
\begin{equation}
  R(A_{ij})\
  = 1+(\alpha -1)A_{ij}
  = \left\{
      \begin{array}{lr}
        \alpha \;\;\; \text{if} \;\; A_{ij} = 1  \\
        1 \;\;\;\;\;\;\;  \text{otherwise}  \\
      \end{array}
    \right.,
  \label{eq:bij}
\end{equation}
where $\alpha \in [0,1]$ is a hyperparameters. If $x_i$ is the neighbor of $x_j$, the distance in \textit{structure space} will be reduced by $\alpha$, and the similarity of $x_i$ and $x_j$ will increase.

\subsection{Against Local Collapse by a Smoother CL Framework}
\label{subsec_structural}
The CL loss in self-supervised contrastive learning (CL) frameworks is
\begin{equation}
  L_{\text{C}}= -\mathbb{E}_{
        \substack{x_i,x_j}
        }\!
       \left[ 
          A_{ij}\log \kappa ( d_{ij}^{z})
        \!+\!
        (1\!-\!A_{ij} )\log (1\!-\!\kappa (d_{ij}^{z}))
        \right],
  \label{eq:contrastive}
\end{equation}
where similarity kernel function $\kappa ( d_{ij}^{z})$ is defined in Eq.(\ref{eq:t_dis}).
The CL is not smooth and can be analogous to bang-bang control~\cite{flugge2015discontinuous} in control systems. Because the learning target of the point pair will switch between $\log \kappa ( d_{ij}^{z})$ and $\log (1\!-\!\kappa (d_{ij}^{z}))$ with the change of $A_{ij}$.

The proposed framework is a smoother CL framework because DLME compromises the learning process and avoids sharp conflicts in gradients.
To compare the difference between the DLME loss and the CL loss, we assume that $g_{\phi}(\cdot)$ satisfies $K$-Lipschitz continuity \cite{gouk_regularisation_2018}, then 
\begin{equation}
  d^z_{ij} = k^* d^y_{ij} , k^* \in [1/K, K],
\end{equation}  
where $k^*$ is a Lipschitz constant. The difference of CL loss and DLME loss is 
\begin{equation}
  \begin{aligned}
    |L_{\text{D}}\!-\! L_\text{c}| = 
    \mathbb{E}_{\substack{x_j, x_j }} 
      \left[
        A_{ij} 
        \!-\!
        \kappa \left((1\!+\!(\alpha\!-\!1)A_{ij}) k^* d^z_{ij} \right)
      \log
      (
        {\frac{1}{\kappa\!(\! d_{ij}^{z}\!)}}
        \!-\!
        1
      )
      \right],
  \end{aligned}
  \label{eq:lemma1}
\end{equation}  
The detailed derivation is provided in Appendix B.  
If $i$ and $j$ are neighbors, $A_{ij}=1$, when $\alpha \to 0$, then $ \alpha k^* d^z_{ij} \!\to\! 0$ and then $1 - \kappa (\alpha k^* d^z_{ij} ) \!\to\! 0$, finally we have the $|L_{\text{D}}\!-\! L_\text{c}| \!\to\! 0$. 
When $\alpha \to 0$, the two losses have the same effect on the samples within each neighbor system. When $\alpha > 0$, the optimal solution of $L_\text{D}$ retain a remainder about the embedding structure $d^z_{ij}$ (in Appendix) which indicates that the DLME loss does not maximize the similarity of the neighborhood as the CL loss, but depends on the current embedding structure.
Eq.(\ref{eq:lemma1}) indicates that the DLME loss is smoother and can preserve the data structure in the embedding space.
When $\alpha > 0$, the DLME loss is a smooth version of the CL loss, which causes a minor collapse of local structures.

Generally, $f_\theta(\cdot)$ explores the structure of the prior manifolds defined by the given data augmentations smoothly, which can model the manifold structure more accurately than previous ML and DML methods.
\subsection{Why DLME Leads to Local Flatness}
\label{sub_flateness}
This section discusses why the DLME loss optimizes the local curvature to be flatter.
Network $f_{\theta}(\cdot)$ maps the data in the input space to the structure space for accurate structural modeling, although it is not guaranteed to obtain locally flat manifolds.
Curling can cause overlap and deformation of the manifold, which can cause degradation of downstream task performance.
To improve the performance in downstream tasks, we need to obtain an embedding space as flat as possible. The simplest linear methods can perform the discriminative tasks (classification, clustering, visualization).

DLME loss can enforce the flatness of the manifold in the embedded space.
Similar to t-SNE, we use the kernel function of the long-tailed t-distribution to transform the distance metric into similarity. 
Further, we apply different `degrees of freedom' parameters $\nu$ in the two latent spaces.
The differences in the degree of freedom $\nu$ form two different kernel functions $\kappa(d, \nu_y)$ and $\kappa(d, \nu_z)$, 
and the difference of kernel functions will make the manifold in the embedding space flatter during the training process.

As described by Eq. (\ref{eq:curvature}), we use the local curvature description of the discrete surface to represent the flatness of the manifold. Next, we theoretically discuss why DLME's loss can minimize the local curvature. 
We use the Push-pull property to describe the action of DLME loss on the embedding space.

\noindent\fbox{
  \parbox{0.95\textwidth}{
    \noindent \textbf{Lemma 1 (Push-pull property)}.
    let 
    $\nu^y>\nu^z$
    and let $d^{z+} = \kappa^{-1}(\kappa(d, \nu^y), \nu^z)$ be the solution of minimizing $L_\text{D}$. 
    Then exists $d_p$ so that $(d^y - d_p) (d^{z+} - d^y) >0$.
  }
}

\begin{figure}[t]
  \centering
  \includegraphics[width=3.5in]{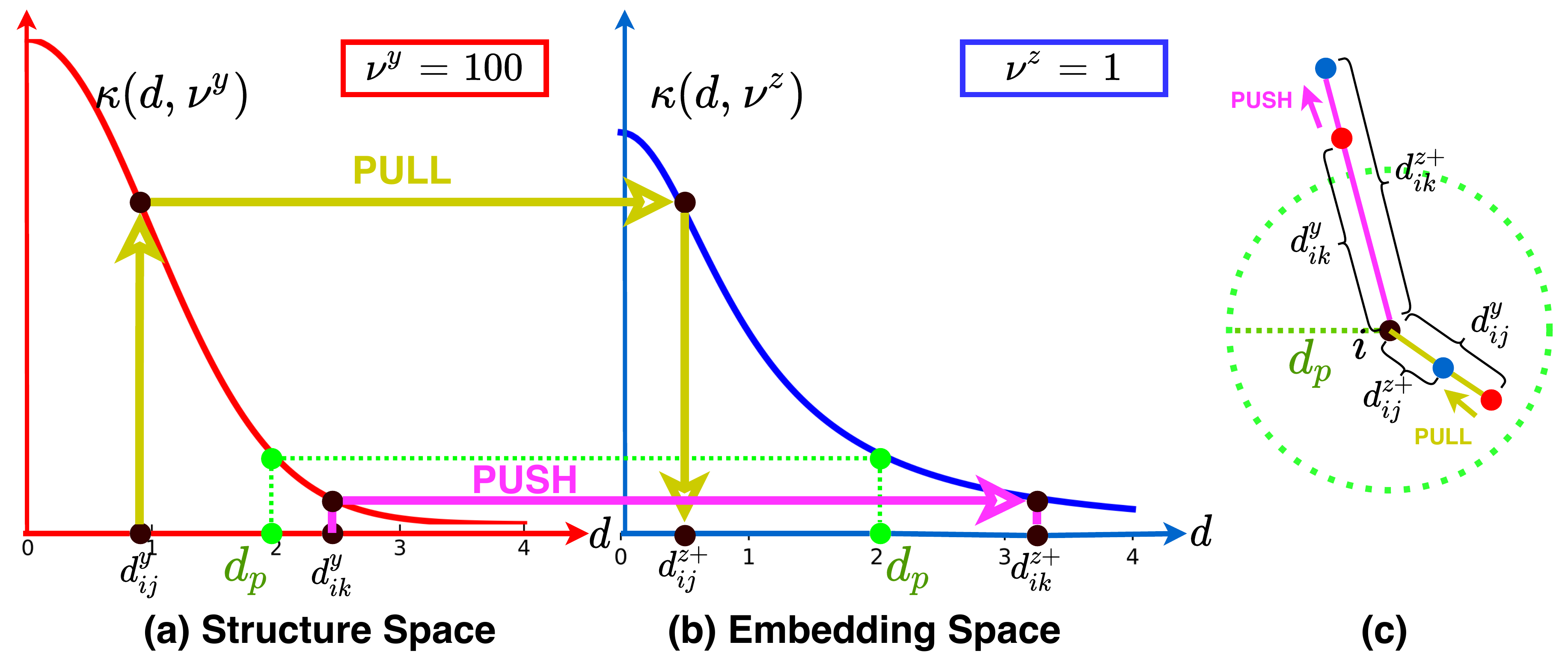}
  \caption{
    Push-pull property: 
    if $d^y<d_p$ then $d^{z+} < d^y$ (in yellow), 
    if $d^y>d_p$ then $d^{z+} > d^y$ (in pink).
    }
  \label{fig:FIg-dl}
\end{figure}

The proof of lemma~1 is detailed in Appendix
. 
Lemma 1 describes the push-pull property of the DLME loss between sample pairs in the embedding space.
$L_\text{D}$ decreases the distance between sample pairs within the threshold $d_p$ (as similar pairs) and increases the distance between sample pairs beyond $d_p$ (as dis-similar pairs), which shows pushing and pulling effects between two kinds of sample pairs. 
Next, we prove that the DLME loss minimizes the average local curvature of the embedding based on the push-pull property.

\noindent\fbox{
  \parbox{0.95\textwidth}{
    \noindent \textbf{Lemma 2}. 
    Assume $f_\theta(\cdot)$ satisfies HOP1-2 order preserving:
      $
      \max ( \{ d_{ij}^y \}_{j \in \text{H}_1(x_i)} ) < \min ( \{ d_{ij}^z \}_{j \in \text{H}_2(x_i)})
      $
    Then $\overline{K}_\mathcal{M}^{z+}<\overline{K}_\mathcal{M}^y$ where $\overline{K}_\mathcal{M}^y$ is the mean curvature in the structure space, and $\overline{K}_\mathcal{M}^{z+}$ is the mean curvature optimization results of $L_\text{D}$ in the embedding space.
  }
}

Lemma 2 indicates that the DLME loss encourages the flatness of the embedding space by decreasing the local average curvature. 
As Fig.~\ref{fig:swishrollandbell}, Lemma 2 describes that the optimization result of DLME loss is to flatten the manifold of the embedded space, which means that we can represent the data in a latent space as linear as possible.
\textbf{DLME's pseudo-code} is shown in Algorithm \ref{alg:algorithm}.
\begin{algorithm}[h]
    \caption{The DLME algorithm}
    \label{alg:algorithm}
    \textbf{Input}: 
        Data: $\mathcal X = \{x_i\}_{i=1}^{|\mathcal X|}$, 
        Learning rate: $\eta$, 
        Epochs: $E$,
        Batch size: $B$,
        $\alpha$, $\nu^{y}$, $\nu^{z}$,
        Network: $f_\theta, g_{\phi}$,
    \textbf{Output}: Graph Embedding: $\{e_i\}_{i=1}^{|\mathcal X|} $.
    
    \begin{algorithmic}[1] 
        \WHILE{{$i=0$; $i<E$; $i$++}}
        \STATE $ X^{+}  \!\leftarrow\! X \cup \tau (X) $.
        \textcolor{green!55!blue}{\# Data augmentation}\\
            \WHILE{{$b=0$; $b<[ |\mathcal X| /B]$; $b$++}}
            \STATE $ \{x_{a,1}, x_{a,2}\!\sim\!X^{+}\!\}_{a\in\mathcal{B}}$, $\mathcal{B} = \{1,\cdots, B\};$
            \textcolor{green!55!blue}{\# Sampling}\\
            \STATE $\{y_{a,0}, y_{a,1} \!\leftarrow\!f_{\theta}(x_{a,0}),f_{\theta}(x_{a,1})\}_{a\in\mathcal{B}}$;
            \textcolor{green!55!blue}{\# Map to $\mathbb{R}^{d_y}$} \\
            \STATE $\{z_{a,0}, z_{a,1} \!\leftarrow\!g_{\phi}(y_{a,0}),g_{\phi}(y_{a,1})\}_{a\in\mathcal{B}}$;
            \textcolor{green!55!blue}{\# Map to $\mathbb{R}^{d_z}$} \\
            \STATE $\{d_{a,ij}^y \!\leftarrow\!d (y_{a,0}, y_{a,1})\}_{a\in\mathcal{B}}$; $\{d_{a,ij}^z \!\leftarrow\!d (z_{a,0}, z_{a,1})\}_{a\in\mathcal{B}}$;
            \textcolor{green!55!blue}{\#Cal. dist in $\mathbb{R}^{d_y}$ \& $\mathbb{R}^{d_z}$} \\
            \STATE $\{S^y_a\!\leftarrow\!\kappa\!(R(B_{a,ij}\!) d^y_{a,ij}\!,\!\nu_y\!)\}_{a\in\mathcal{B}}$; $\{S^z_a \!\leftarrow\! \kappa (d^z_{a,ij}, \nu_z )\}_{a\in\mathcal{B}}$; 
            \textcolor{green!55!blue}{\#Cal. sim in $\mathbb{R}^{d_y}$ \& $\mathbb{R}^{d_z}$} \\
            \STATE $\mathcal{L}_{\text{D}} \!\leftarrow\! E( \{D(S^y_a, S^z_a)\}_{a\in\mathcal{B}})$ by Eq.~(\ref{eq:final_model});
            \textcolor{green!55!blue}{\# Cal. loss function} \\
            \STATE $\theta \!\leftarrow\!  \theta - \eta \frac{ \partial \mathcal{L}_{\text{D}} }{\partial \theta}$, $\phi \!\leftarrow\! \phi - \eta \frac{ \partial \mathcal{L}_{\text{D}} }{\partial \phi}$;
            \textcolor{green!55!blue}{\# Update parameters} \\
            \ENDWHILE
        \ENDWHILE
        \STATE $ \{z_i \!\leftarrow\! f_{\theta}(g_{\phi}(x_i))\}_{i \in \{ 1,2,\cdots,\mathcal X \}}$;
        \textcolor{green!55!blue}{\# Cal. the embedding result} \\
    \end{algorithmic}
\end{algorithm}

\section{Experiments}
\label{sec_exp}

In this section, we evaluate the effectiveness of the proposed DLME on four downstream tasks (classification/linear test, clustering, visualization) and analyze each proposed component with the following questions.\\
($\bf{Q}$1) How to intuitively understand structural distortions?
($\bf{Q}$2) Does DLME overcome structural distortions?
($\bf{Q}$3) How to intuitively understand underconstrained manifold embedding?
($\bf{Q}$4) Does DLME overcome underconstrained manifold embedding and obtain locally flat embeddings?
($\bf{Q}$5) How much does DLME improve the performance of downstream tasks on ML and CL datasets?
($\bf{Q}$6) Can smoother losses bring better performance to CL?
\subsection{Visualization of Structural Distortions~($\bf{Q}$1,$\bf{Q}$2)}
\textbf{Experimental setups.} 
This section illustrates structural distortions on image datasets and experimentally demonstrates that DLME can overcome structural distortions by introducing prior knowledge of data augmentation.
In this experiment, all the data are mapped to a 2-D latent space to facilitate the visualization. All compared ML methods (t-SNE, PUMAP, ivis, and PHA) will fail in the CIFAR dataset; we only show the results of UMAP.

\textbf{Structural Distortions.} 
The ML approach uses distance metrics from observations to model the structure. The complexity of the data (data with dimensionality and not-well sampling) leads to a failure of the distance metric, confusing the semantic nearest neighbors and subsequently destroying local connectivity, ultimately creating structural distortions in the ML process.
DLME constructs a smoother CL framework with the help of data augmentation. The proposed framework obtains richer prior knowledge with data augmentation. It maps the data into the latent space for structural modeling with neural networks, which can achieve more accurate modeling and thus overcome structural distortion.

\begin{figure}[t]
    \centering
    \includegraphics[width=4.7in]{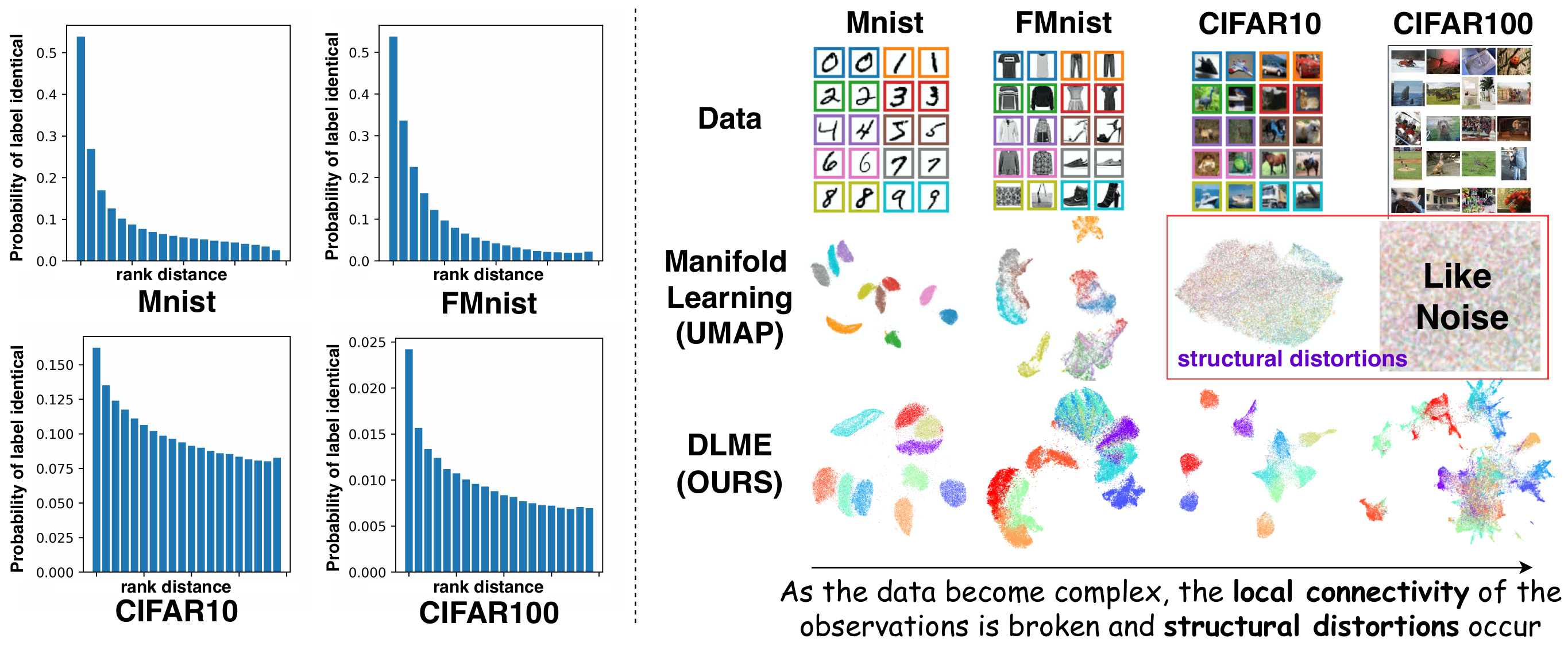}
    \caption{
        (left) The Bar plot of probabilities of identical label v.s. rank distance. A higher left end of the bar plot indicates a higher probability of the same label for the nearest neighbor sample, implying that local connectivity is guaranteed. (right) The results of ML methods (UMAP) for four image datasets. For complex data, local connectivity cannot be guaranteed, leading to the embedding failure of the ML method. The proposed DLME method has better embedding results on the more complex CIFAR dataset.
    }
    \label{fig:structural_distortions}
\end{figure}
\subsection{Visualization of Underconstrained Manifold Embedding ($\bf{Q}$3,$\bf{Q}$4)}

This section illustrates underconstrained manifold embedding with toy datasets and experimentally demonstrates the DLME potential to solve this problem by constraining the local curvature.
\begin{figure}[t]
    \centering
    \includegraphics[width=4.7in]{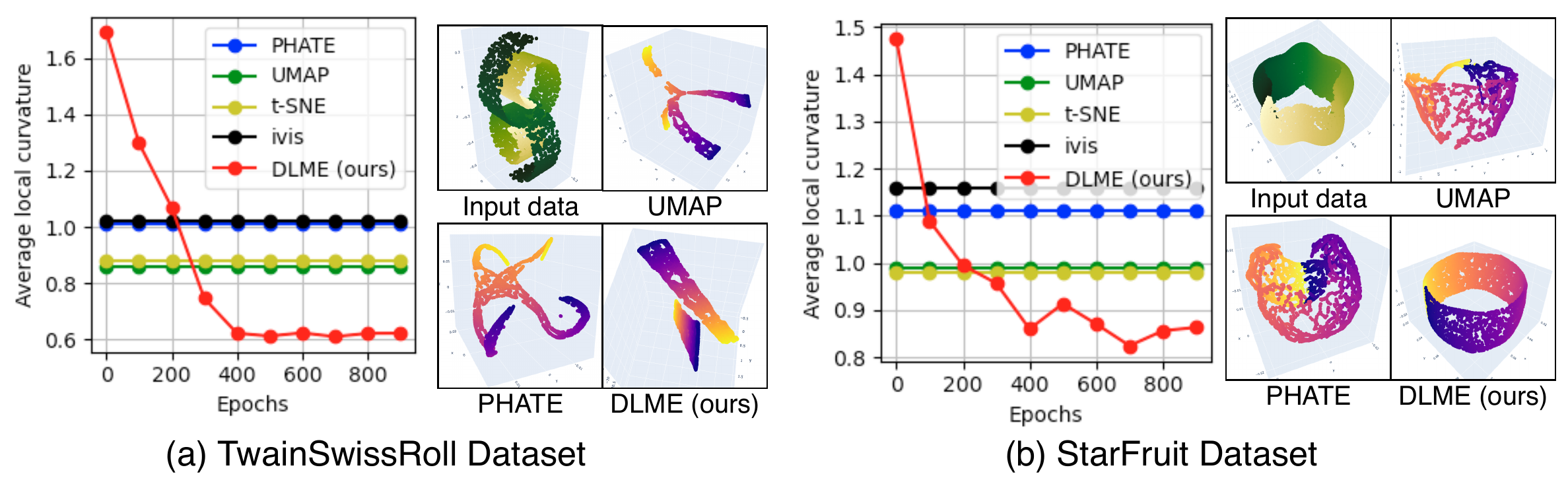}
    \caption{
        Average local curvature and scatter plot on TwainSwissRoll and StarFruit dataset. 
        The two examples indicate that traditional ML produces distorted embeddings, which affect the performance of downstream tasks. In contrast, DLME can get as flat embeddings as possible by optimizing the local curvature.
    }
    \label{fig:swishrollandbell}
\end{figure}
\textbf{Experimental setups.} 
The experiments include two 3D toy datasets, TwainSwissRoll and StarFruit. The TwainSwissRoll dataset has two tangled but disjoint SwissRolls. The StarFruit dataset has a locally curved surface. The input data and outputs of compared methods are shown in Fig.~\ref{fig:swishrollandbell}. The details of the datasets and outputs are shown in Appendix.

\textbf{Underconstrained manifold embedding.} 
Fig.~\ref{fig:swishrollandbell} shows that the compared ML methods produce bends that should not exist. We attribute these bends to inadequate constraints on the loss function. These bends affect downstream tasks such as classification and clustering. In addition, these bends may cause more significant damage when the data situation is more complex.

\textbf{Benefits of local flat embedding.}
Since a flat localization is assumed, DLME tries to obtain a locally flat embedding. The statistics of the mean curvature show that DLME can receive a flatter local manifold. We consider that a flat embedding possesses practical benefits. A flatter local manifold can achieve better performance in downstream tasks and is more suitable for interpretable analysis of deep models. For example, it is easy to distinguish two sub-manifolds of TwainSwissRoll using a linear classifier, and it is easy to perform regression tasks on StarFruit.
\subsection{Comparison on Traditional ML Datasets ($\bf{Q}$5, $\bf{Q}$6)}
\label{sub_easy}
\textbf{Experimental setups:}
The compared methods include two ML methods (UMAP~\cite{mcinnes_umap_2018}, t-SNE~(tSNE)~\cite{hinton_reducing_2006}) and three deep ML methods (PHATE~(PHA) \cite{moon2019visualizing_}, ivis \cite{szubert_structure_preserving_2019} and parametric UMAP~(PUM) \cite{sainburg_parametric_2021}.) The experiments are on six image datasets~(Digits, Coil20, Coil100, Mnist, EMnist, and KMnist) and six biological datasets (Colon, Activity~(Acti), MCA, Gast, SAMUSIK~(SAMU), and HCL). For a fair comparison, all the compared methods map the input data into a 2D space and then evaluated by 10-fold cross-validation. The MLP architecture of $f_{\theta}$ is [-1,500,300,80], where -1 is the dimension of input data. The MLP architecture of $g_{\phi}$ is [80,500,80,2]. The comparison results of linear SVM and k-means are shown in Table~\ref{tab:SimpleManifold}. Details of datasets, baseline methods, and evaluation metrics are in Appendix.

\begin{table}[]
    \caption{Performance comparison on 12 datasets. \textbf{Bold} denotes the best result and \underline{\textbf{Underline}} denotes 5\% higher than others.}
    \centering
    \begin{tabular}{@{}c|ccccc|c|ccccc|c@{}}
    \toprule
            & \multicolumn{6}{c|}{Classification Accuracy (linear SVM)}                      & \multicolumn{6}{c}{Clustering Accuracy (K-means)}                       \\ \cmidrule(l){2-13}
            & tSNE           & UMAP  & PUM     & ivis  & PHA   & DLME                                  & tSNE  & UMAP   & PUM    & ivis  & PHA & DLME                          \\ \midrule
    Digits  & 0.949          & 0.960 & 0.837   & 0.767 & 0.928 & \textbf{0.973}                        & 0.938 & 0.875  & 0.763  & 0.726 & 0.794 & \textbf{0.956}                \\
    Coil20  & 0.799          & 0.834 & 0.774   & 0.672 & 0.828 & \underline{\textbf{0.909}}            & 0.763 & 0.821  & 0.722  & 0.612 & 0.655 & \underline{\textbf{0.899}}    \\
    Coil100 & 0.760          & 0.756 & N/A     & 0.542 & 0.653 & \underline{\textbf{0.952}}            & 0.763 & 0.785  & N/A    & 0.492 & 0.515 & \underline{\textbf{0.944}}    \\
    Mnist   & 0.963          & 0.966 & 0.941   & 0.671 & 0.796 & \textbf{0.976}                        & 0.904 & 0.801  & 0.772  & 0.466 & 0.614 & \underline{\textbf{0.977}}    \\
    EMnist  & 0.420          & 0.588 & 0.384   & 0.190 & 0.416 & \underline{\textbf{0.657}}            & 0.478 & 0.537  & 0.363  & 0.178 & 0.352 & \underline{\textbf{0.641}}    \\
    KMnist  & 0.738          & 0.656 & 0.674   & 0.547 & 0.607 & \underline{\textbf{0.782}}            & 0.586 & 0.668  & 0.706  & 0.522 & 0.594 & \textbf{0.712}                \\ \midrule
    Colon   & 0.932          & 0.893 & 0.918   & 0.942 & 0.930 & \textbf{0.947}                        & 0.862 & 0.847  & 0.861  & 0.922 & 0.855 & \textbf{0.924}                \\
    Acti    & 0.861          & 0.844 & 0.849   & 0.831 & 0.798 & \underline{\textbf{0.921}}            & 0.784 & 0.639  & 0.783  & 0.681 & 0.679 & \underline{\textbf{0.898}}    \\
    MCA     & 0.719          & 0.675 & 0.667   & 0.634 & 0.552 & \underline{\textbf{0.774}}            & 0.475 & 0.532  & 0.464  & 0.443 & 0.414 & \textbf{0.563}                \\
    Gast    & 0.821          & 0.846 & 0.706   & 0.687 & 0.676 & \underline{\textbf{0.918}}            & 0.534 & 0.546  & 0.512  & 0.427 & 0.523 & \underline{\textbf{0.598}}    \\
    SAMU    & 0.556          & 0.678 & 0.599   & 0.625 & 0.675 & \textbf{0.700}                        & 0.335 & 0.387  & 0.345  & 0.328 & 0.511 & \underline{\textbf{0.572}}                \\
    HCL     & \textbf{0.874} & 0.863 & 0.767   & 0.454 & 0.393 & \textbf{0.874}                               & 0.689 & 0.743  & 0.619  & 0.308 & 0.263 & \textbf{0.753}                \\
    \bottomrule
    \end{tabular}
    \label{tab:SimpleManifold}
\end{table}

\textbf{Analyse:}
DLME has advantages over all 12 datasets, and DLME is 5\% higher than other methods in 14 items (24 items in total).
We observe that the proposed DLME has advantages in classification and clustering metrics. We summarize the reasons why DLME has performance advantages as follows. (1) Compared with ML methods, DLME overcomes structural distortions to a certain extent to model the structure more accurately. (2) DLME reduces the overlap of different clusters, improving the performance of classification and clustering. (3) The locally flat embeddings learned by DLME are linearly characterized and more suitable for the linear model.
\subsection{Comparison on CL Datasets ($\bf{Q}$5, $\bf{Q}$6)}
\begin{table}[]
    \centering
    \caption{The linear-test Performance comparison on image datasets.}
    \begin{tabular}{@{}ccccccccccc@{}}
        \toprule
        Dataset       & \multicolumn{1}{c}{CIFAR10}   & \multicolumn{1}{c}{CIFAR100} & \multicolumn{1}{c}{STL10} & \multicolumn{1}{c}{TinyImageNet}  & \multicolumn{1}{c}{ImageNet100} \\ \midrule
        NPID          & 0.827                      & 0.571                     & 0.825                     & 0.382                      & 0.721                       \\
        ODC           & 0.799                      & 0.521                     & 0.734                     & 0.287                      & 0.645                       \\
        SimCLR        & 0.882                      & 0.574                     & 0.869                     & 0.384                      & 0.756                       \\
        MoCo.v2       & 0.886                      & 0.614                     & 0.856                     & 0.374                      & 0.780                       \\
        BYOL          & 0.881                      & 0.644                     & 0.887                     & 0.388                      & 0.785                       \\ \midrule
        {\bf DLME}    & \underline{\textbf{0.913}} & {\textbf{0.661}}          & \textbf{0.901}            & \underline{\textbf{0.449}} & {\textbf{0.793}}            \\
        {\bf DLME-A1} & {0.910}                    & {0.653}                   & {0.881}                   & {0.428}                    & {0.785}                     \\
        {\bf DLME-A2} & {0.902}                    & {0.626}                   & {0.879}                   & {0.432}                    & {0.791}                     \\
        {\bf DLME-A3} & {0.888}                    & {0.624}                   & {0.873}                   & {0.401}                    & {0.783}                     \\ \bottomrule
    \end{tabular}
    \label{tab:leneartest}
\end{table}

\begin{table}[]
    \centering
    \caption{The clustering Performance comparison on image datasets.}
    \begin{tabular}{@{}cccccccccccc@{}}
        \toprule
        Dataset       & \multicolumn{1}{c}{CIFAR10} & \multicolumn{1}{c}{CIFAR100} & \multicolumn{1}{c}{STL10} & \multicolumn{1}{c}{TinyImageNet} & \multicolumn{1}{c}{ImageNet-Dog} \\ \midrule
        DAC           & 0.522                    & 0.238                     & 0.470                     & 0.066                   & 0.219                   \\
        DCCM          & 0.623                    & 0.327                     & 0.482                     & 0.108                   & 0.321                   \\
        PICA          & 0.696                    & 0.337                     & 0.713                     & 0.098                   & 0.352                   \\
        CC            & 0.747                    & 0.429                     & 0.850                     & 0.140                   & 0.445                   \\
        CRLC          & 0.799                    & 0.425                     & 0.818                     & 0.153                   & 0.461                   \\ \midrule
        {\bf DLME}    & \underline{{\bf 0.822}}  & {\bf 0.441}               & \underline{{\bf 0.883}}   & \underline{{\bf 0.182}} & \underline{{\bf 0.483}} \\
        {\bf DLME-A1} & {0.792}                  & {0.417}                   & {0.872}                   & {0.145}                 & {0.479}                 \\
        {\bf DLME-A2} & {0.783}                  & {0.421}                   & {0.859}                   & {0.133}                 & {0.480}                 \\
        {\bf DLME-A3} & {0.779}                  & {0.417}                   & {0.852}                   & {0.134}                 & {0.477}                 \\ \bottomrule
    \end{tabular}
    \label{tab:imageclusting}
\end{table}

\begin{figure*}[t]
    \centering
    \includegraphics[width=4.7in]{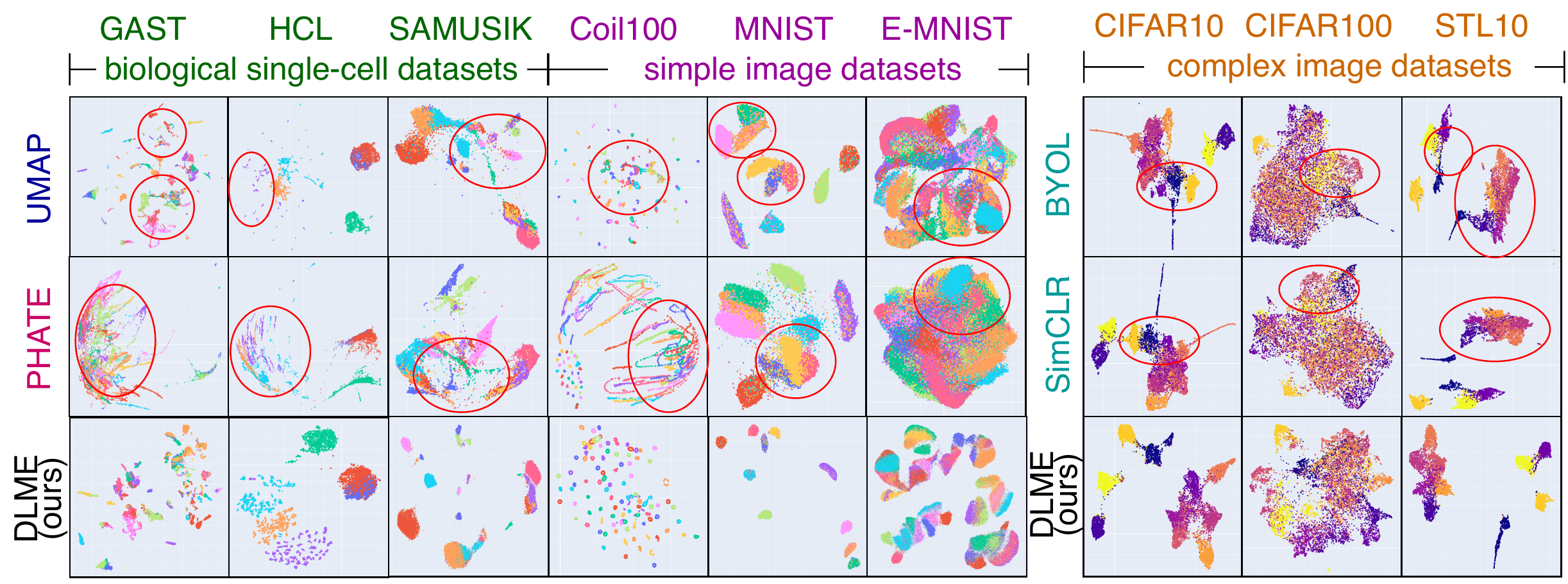}
    \caption{
    Visualization results of biological datasets, simple image datasets and complex image datasets.
    The red circle indicates the clusters confused by the baseline method. 
    }
    \label{fig:FIg-vis}
\end{figure*}

\textbf{Experimental setups:}
Due to structural distortions, ML methods fail in CV Datasets, and our comparison is limited to the CL and deep clustering (DC) domain.
The compared methods include CL methods (
    NPID \cite{wu_unsupervised_2018}, 
    ODC \cite{zhan2020online}, 
    SimCLR \cite{chen_simple_2020}, 
    MOCO.v2 \cite{he_momentum_2020} and 
    BYOL \cite{grill_bootstrap_2020}
)
and deep clustering methods (
    DAC~\cite{haeusser_associative_2019}, 
    DDC \cite{chang_deep_2019}, 
    DCCM \cite{wu_deep_2019-1}, 
    PICA \cite{huang_deep_2020}, 
    CC \cite{li_contrastive_2020}, and
    CRLC \cite{do2021clustering}
).
The datasets include six image datasets: CIFAR10, CIFAR100, STL10, TinyImageNet, ImageNet-Dog and ImageNet100.
The $f_{\theta}$ is ResNet50, and $g_{\phi}$ is MLP with architecture of [2048, 256].
We use the same settings as SimCLR for the linear test and use the same settings as CC~\cite{li_contrastive_2020} for deep clustering.
The results are shown in Table~\ref{tab:leneartest} and Table~\ref{tab:imageclusting} and the detailed setup is in Appendix
.

\textbf{Analyse:}
In all datasets, DLME outperformed the SOTA method by a large margin. And it beat the other techniques by 2\% in 6 items (out of 10 items). The reason is the smoother DLME framework avoids problems such as falling into local collapse. Another reason is locally flat embeddings learned by DLME are linearly characterized and more suitable for a linear model.

\textbf{Ablation Study.}
We designed ablation experiments to demonstrate the effectiveness of DLME. \textbf{Ablation 1 (DLME-A1)}, we detach the $L_\text{D}$'s gradient on the model $f_\theta(\cdot)$ and replace it with the CL loss (in Eq.(\ref{eq:contrastive})). The model is divided into two separate parts. One obtains embedding with CL, and the other emphasizes the flatness of manifold with similarity loss. \textbf{Ablation 2 (DLME-A2)}, based on DLME-A1, We ablate the t-distribution kernel and use a standard distribution kernel in both spaces. \textbf{Ablation 3 (DLME-A3)}, finally, we further ablate the structure of the two networks and transform the model into a CL method. The results of ablation experiments are in Table \ref{tab:leneartest} and Table \ref{tab:imageclusting}. The experimental results show that the three critical operations of DLME can improve the performance in complex manifold embedding tasks.
\subsection{Visualization of ML and CV Datasets ($\bf{Q}$5, $\bf{Q}$6)}
DLME is an appropriate method for visualizing high-dimensional data. A typical setup for data visualization using DLME is to embed the data directly into 2D space. As the selected visualization results are shown in Fig.~\ref{fig:FIg-vis}, DLME significantly outperforms other methods in terms of visualization results. Because the distortion problem is overcome, the DLME embedding results in a minimum mixture of different clusters with clear boundaries. The detailed results are shown in Appendix.

\section{Conclusion}
We propose Deep Local-flatness Manifold Embedding (DLME), a novel ML framework to obtain reliable manifold embedding by reducing distortion.
In the experiments, by demonstrating the effectiveness of DLME on downstream classification, clustering, and visualization tasks with three types of datasets (toy, biological, and image), our experimental results show that DLME outperforms SOTA ML \& CL methods.

\section*{Acknowledgement}
This work is supported by National Natural Science Foundation of China, named Geometric Deep Learning and Applications in Proteomics-Based Cancer Diagnosis (No. U21A20427). This work is supported by Alibaba Innovative Research (AIR) Programme. 
\newpage

\clearpage
%
%
\bibliographystyle{splncs04}
\bibliography{MyLibrary}

\end{document}